\documentclass[11pt,a4]{article}
\usepackage{coling2016}
\usepackage{acl-citealt}
\usepackage{url}
\usepackage{latexsym}
\usepackage{booktabs}
\usepackage{microtype}
\usepackage{paralist}

\title{Minimally Supervised Written-to-Spoken Text Normalization}

\author{Ke Wu \and Kyle Gorman \and Richard Sproat\\
  {\tt \{wuke,kbg,rws\}@google.com}}
\date{}

\begin{document}

\maketitle

\begin{abstract}

In speech-applications such as text-to-speech (TTS) or automatic speech
recognition (ASR), \emph{text normalization} refers to the task of
converting from a \emph{written} representation into a representation of how the
text is to be \emph{spoken}.
In all real-world speech applications, the text normalization engine is developed---in large part---by hand.
For example, a hand-built grammar may be used to enumerate the possible ways of saying a given token in a given language, and a statistical model used to select the most appropriate pronunciation in context.
In this study we examine the tradeoffs associated with using more or less language-specific domain knowledge in a text normalization engine.
In the most data-rich scenario, we have access to a carefully
constructed hand-built normalization grammar that for any given token will
produce a set of all possible verbalizations for that token. We also assume a
corpus of aligned written-spoken utterances, from which we can train a ranking model that selects the appropriate verbalization for the given context.
As a substitute for the carefully constructed grammar, we also consider a scenario with a language-universal normalization \emph{covering grammar}, where the developer merely needs to provide a set of lexical items particular to the language. As a
substitute for the aligned corpus, we also consider a scenario where one only has the spoken side,
%(or only ``clean'' written text),
and the corresponding written side is ``hallucinated'' by composing the spoken side with the inverted normalization grammar.
We investigate the accuracy of a text normalization engine under each of these scenarios.
%This question is of direct relevance to low-resource languages, where it may be
%possible to get raw data and some lexical information, but investing the effort
%to develop high-quality grammars and training data may not be practical.
We report the results of experiments on English and Russian.
%., and on the highly-inflected language Russian.
% I have killed "highly-infleceted" because I think either people will already know that Russian is difficult for that reason, or they won't know what "inflection" is. Besides, it seems to foreground information that doesn't seem within the scope of an abstract. --kbg. 

\end{abstract}

\section{Introduction}

Over the past 15 years there has been substantial progress on machine-learning approaches to text normalization \cite{Sproat2001,Schwarm2002,Han2011,Liu2011,Pennell2011b,Yang2013b}. Nonetheless,
for real-world speech applications, such as text-to-speech (TTS) systems or automatic speech recognition systems (ASR), text normalization engines require a substantial amount of manual grammar development.
Despite powerful machine-learning methods, it is still not possible to learn accurate end-to-end text normalization engines for speech applications using only a large amount of annotated data.

One reason for this is a lack of appropriate annotated data.
Constructing hand-built grammars is a labor intensive and time consuming process, but so is constructing appropriate corpora consisting of raw text and the corresponding normalization.\footnote{Minimally supervised approaches that attempt to construct normalizers using \emph{unannotated} text data based on matching unnormalized forms with possible expansions given some corpus of `clean' text \cite{Sproat2001,Yang2013b,Roark2014} are relevant, but are only applicable to certain classes of tokens, such as abbreviations.}
Furthermore training corpora for text normalization must be constructed for the particular application they are intended for.
There are, for example, several training corpora for social media text normalization \cite{Han2011,Liu2011,Yang2013b}, but these are of little use for a TTS system where the normal input rarely includes examples like \emph{cu l8tr lv u ;-)} but where, on the other hand, numbers need to be expanded into words (e.g., \emph{twenty-three thousand}), something that is generally of no interest when normalizing social media text.

In this paper we present a collection of methods to learn a mapping from \emph{written} text to its \emph{spoken} form, i.e., the mapping needed for a TTS system, or for producing language model training data from raw text.
For our data we use 2.24 million words of English transcribed speech and 1.95 million words of Russian speech, both collected from a voice search application.
This \emph{spoken} form data was converted to plausible \emph{written} forms by a semiautomatic process, then corrected by a team of native-speaker annotators.  This results in 1.75 million written tokens for English, and 1.59 million written tokens for Russian.\footnote{Any personally identifiable information was removed from this data before analysis.}

In some of our studies we also assume a hand-built language-specific normalization grammar, henceforth referred to as the \textbf{language-specific (normalization) grammar}.
In this study, we use pre-existing grammars developed as part of a system for training language models for ASR.
The normalization grammars are represented as weighted finite-state transducers (WFST) compiled using the Thrax grammar development system \cite{Thrax}.
This grammar is designed to \emph{overgenerate}; that is, it produces, for any input token, a (potentially weighted) lattice of possible spoken strings --- see \cite{Gorman:Sproat:16}.
Thus for example \emph{123} might produce \emph{one hundred twenty three}, \emph{one two three}, \emph{one twenty three}, among possibly others.
%The correct variant for a given context is assumed to be selected by a language model or in any case some model that uses features of the context.
In Section~\ref{sec:ranking} we describe how we combine this grammar with a reranker that selects the appropriate choice in context.
These language-specific grammars are written by native speaker linguists and designed to cover a range of types of normalization phenomena, including numbers, some mathematical expressions, some abbreviations, times and currency amounts.
Following \newcite{Taylor2009}, we refer to these categories targeted by normalization as \emph{semiotic classes}.

Such grammars, needless to say, take a fair amount of effort and knowledge to develop for a new language.
But in fact, there is a lot of information that can be shared across languages if the grammars are appropriately parameterized.
To take a simple example, a time such as \emph{3:30 PM} might be read as the equivalent of \emph{three thirty PM} as in English, or as \emph{PM three thirty}, with the period expression before the time, as in Chinese, Japanese or Korean. It is thus possible to construct what we shall henceforth refer to as a \textbf{covering grammar}, which sets out the universally available options.
For \emph{3:30 PM} example, the covering grammar would allow verbalizations in both orders.
All that is required to specialize the covering grammar for a specific language is to provide a small lexicon providing verbalizations for individual terms (e.g., how \emph{PM} is read): such a lexicon could be as small as a few hundred items, and can be developed by a native speaker who is not a trained linguist.
Something more than this is required for readings of numbers, which are considerably less constrained: for that we make use of the algorithm described in \cite{Gorman:Sproat:16}, which can learn verbalizations for number names in a language from about 300 labeled examples.
Given the covering grammar and the language-specific lexicon, we leave it up to the ranker to learn whether, e.g., one reads \emph{PM} before or after the \emph{3:30}.
In Section~\ref{sec:ranking}, we describe the use of a covering grammar in place of the more-carefully engineered language-specific grammar considered, and consider the degradation resulting from replacing the language-specific grammar with the covering grammar.
Thus we have two kinds of \emph{normalizers}: one derived from a hand-constructed language-specific grammar, and the other from a permissive language-universal covering grammar composed with a language-specific lexicon.

Finally, we consider one other scenario, one where we have only the \emph{spoken} data, bu have not gone to the effort to create the corresponding written forms at all.  This scenario may obtain when one has transcriptions of speech in some language, or a corpus of fairly clean written text where even things like numbers are written out in words. This is a situation that is likely to obtain in low-resource languages where one might not have the resources to create large amounts of parallel data. Our approach in this scenario is to ``hallucinate'' the written-side text by \emph{inverting} the normalizer to turn it into a \emph{denormalizer} \cite{Shugrina2010,Vasserman2015}, and then composing the resulting lattice of putative written forms with the normalizer.\footnote{
Inversion of the verbalizer is trivial if one represents the grammatical knowledge in terms of a WFST---but much harder if it is represented, for example, as a neural network.}
We discuss this process in Section~\ref{sec:ranking}.
Note that this final scenario has much in common with prior work by \citealt{Sagae:EtAl:12} and \citealt{Smith:Eisner:05}.

The above scenarios yield four training conditions, as follows:

\begin{compactenum}
\item language-specific grammar with real written data
\item language-specific grammar with hallucinated written data
\item covering grammar with real written data
\item covering grammar with hallucinated written data
\end{compactenum}
\noindent
As one might expect, the first condition yields the best results and, usually the fourth condition the worst results. But our primary question is how much of a loss in accuracy results from the lower-resource conditions, and we report in detail on this below.
We perform experiments both on English and Russian.
In general the English results are overall much better than those of Russian
---not surprisingly given the complex morphology of the latter---but the results for these two languages are otherwise comparable.

This work thus takes a somewhat different approach than many of the recent machine-learning approaches to text normalization, work best exemplified by research on social media normalization (e.g., \citealt{Yang2013b}).
In particular we are prepared to assume \emph{some} amount of linguistic resource development, including the development of language-specific lexical information, the possible existence of aligned written-spoken data, and for numbers, a minimal set of digit-sequence-to-number-name mappings as required by the method in \cite{Gorman:Sproat:16}.
For readings of numbers, methods such as those of \citealt{Yang2013b} are in any case completely inapplicable. Instead we focus on the relationship between the  amount of language-specific effort required and accuracy of the resulting system.

\section{Data and Conventions}
\label{sec:data}

\begin{table*}
  %\label{tab:corpus_en}
  \label{tab:corpus}
  \centering
  \begin{tabular}{lllll}
  \toprule
  Subset      & \# of sentences & \# of written tokens & \# of spoken tokens & \# of unchanged tokens                     \\
  \midrule
\\
  \textbf{English} \\
\cmidrule{1-1}
  Training    & 325 thousand    & 1.75 million         & 2.24 million        & 1.38 million                               \\
  Development & 5,000           & 27,602               & 34,899              & 21,743                                     \\
  Testing     & 5,000           & 28,225               & 35,645              & 22,345                                     \\
  %\end{tabular}
  %\caption{Corpus size statistics for English}
%\end{table*}
%
%\begin{table*}
%  \label{tab:corpus_ru}
%  \centering
%  \begin{tabular}{lllll}
%  \toprule
%  Subset      & \# of sentences & \# of written tokens & \# of spoken tokens & \# of unchanged tokens                     \\
%  \midrule
\\
\textbf{Russian} \\
\cmidrule{1-1}
  Training    & 348 thousand    & 1.59 million         & 1.95 million        & 1.15 million                               \\
  Development & 2,000           & 9,291                & 11,305              & 6,755                                      \\
  Testing     & 2,000           & 8,874                & 11,002              & 6,376                                      \\
  \bottomrule
  \end{tabular}
  \caption{Corpus size statistics for English and Russian.}
%  \caption{Corpus size statistics for Russian.}
\end{table*}

As noted above, our data consists English and Russian transcribed speech, both collected from a voice search application.
This \emph{written} form data has been converted to plausible \emph{spoken} forms by a semiautomatic process, then corrected by native-speaker annotators.
%% TODO: is this true?
Each sentence in both corpora contains at least one token that requires normalization.
Disjoint subsets are held-out from the corpora as development and test sets.
Table \ref{tab:corpus} lists some basic statistics of the English and the Russian corpora.
As one can see, the majority of the tokens remain unchanged after normalization.
% TODO(rws): We must say something about the data being anonymized.

The parallel data is assumed to be aligned at the word level, so that each written token aligns to one or more spoken tokens. % TODO(wuke): Describe how we get this if there is space.
Therefore we represent a training sentence as a sequence of pairs $(x_1, \mathbf{z}_1), \ldots, (x_n, \mathbf{z}_n)$, where $x_i$ is a written token and $\mathbf{z}_i$ is the sequence of spoken tokens it aligns to. Note that here and below, we use lowercase letters (e.g., $x$) to represent words and scalars and bold lowercase letters (e.g., $\mathbf{y}$) for sequences of words and for vectors.
For any operation that takes a string operand (e.g., composition), we consider a word sequence $\mathbf{y}$ to be synonymous to the string constructed by joining the elements of $\mathbf{y}$ with space.
Figure \ref{fig:aligned-sentence} is an example of an annotated parallel sentence.

\begin{figure}
\begin{center}
\begin{tabular}{ll}
$x_1$ = \texttt{wake} & $\mathbf{z}_1$ = \texttt{wake} \\
$x_2$ = \texttt{me} & $\mathbf{z}_2$ = \texttt{me} \\
$x_3$ = \texttt{at} & $\mathbf{z}_3$ = \texttt{at} \\
$x_4$ = \texttt{9:00} & $\mathbf{z}_4$ = \texttt{nine} \\
$x_5$ = \texttt{am} & $\mathbf{z}_{5,6}$ = \texttt{a m} \\
\end{tabular}
\end{center}
\caption{\label{fig:aligned-sentence} An example word-aligned sentence.}
\end{figure}

As discussed above, we assume two types of grammar, namely a language-specific normalization grammar and a covering grammar. The language-specific normalization grammar for English consists of about 2,500 lines of rules and lexical specifications written using the Thrax finite-state grammar development toolkit \cite{Thrax}, and the equivalent Russian grammar has about 4,100 lines.%\footnote{
%  Note that these line counts include rules for the readings of cardinal and ordinal numbers, though we do not use this component of the grammars here.}

The covering grammar is fundamentally language-independent and consists of about 700 lines of Thrax code. In addition we have 75 lexical specifications for English and 220 for Russian, with the bulk of the difference being due to the need to list various potential inflected forms in Russian. This does not however, include the verbalizations for cardinal and ordinal numbers, which, as we have already noted, in the covering grammar setting are learned from a small set of aligned examples using the method described in \cite{Gorman:Sproat:16}.
% [rws] I find the following sentence cryptic (even though I know what it is trying to say.
%These number grammars use the intersection of a covering grammar, language-specific number rules induced from a small amount of labeled training data, and a lexical map.
As described above, the covering grammar is intended to represent all possible ways in which a language might express particular semiotic classes. By way of illustration, consider the grammar fragment below as it is written in Thrax:

\begin{verbatim}
period = @@TIME_AM@@ | @@TIME_PM@@;
space = " " | ("" : " ");
time = Optimize[(period space)? time_variants |
                time_variants (space period)?];
\end{verbatim}

% Why is this inline but the lexical map is a float? --kbg
% Dunno, why not?

\noindent
The first expression defines  \verb|period| to be either \verb|@@TIME_AM@@| or \verb|@@TIME_PM@@|, which will be verbalized with appropriate lexical items in the target language.
The third expression then defines a time to be any of a number of ways to  verbalize core time values---e.g., \emph{three thirty} or \emph{half past three}, termed \verb|time_variants| here---either preceded or followed by an optional \verb|period| and an intervening space.
For any target language, this will \emph{overgenerate}, so we require a means of selecting the appropriate form for the language; see Section~\ref{sec:ranking}.
A fragment of the lexical map for English is given in Figure~\ref{fig:lex_map}, where variables marked with \verb|@@| are the abstract elements of the grammar that need to be specified.

\begin{figure}
\begin{center}
\begin{tabular}{ll}
\verb|@@OCTOBER@@|  &   \verb|october|\\
\verb|@@NOVEMBER@@| &   \verb|november|\\
\verb|@@DECEMBER@@| &   \verb|december|\\
\verb|@@MINUS@@|    &   \verb|minus|\\
\verb|@@DECIMAL_DOT@@|   &  \verb|point|\\
 \verb|@@URL_DOT@@| & \verb|dot|\\
\end{tabular}
\end{center}
\caption{\label{fig:lex_map}A fragment of the English lexical map.}
\end{figure}

In developing the covering grammar we endeavored to cover the same phenomena as covered in the language-specific verbalization grammars. For example the English verbalization grammar handles various ways of reading digit sequences, so that \emph{990} might be read as \emph{nine hundred ninety}, or \emph{nine ninety} or \emph{nine nine oh}, all of which are possible in various contexts.
In a similar fashion we also allowed digit sequences in the covering grammars to be read as cardinal (or ordinal numbers), or as various combinations of these two categories.
Obviously a fair amount of genre-specific knowledge goes into the design of the covering grammar, and without this knowledge, this strategy would likely fail.
On the other hand, once such a grammar is created, developing a system for a new language merely requires one to specify a lexical map, which requires far less work and expertise than developing a new grammar.\footnote{
	In the hopes that they will be more widely useful, we will release the covering grammars and lexical maps for English and Russian under the Apache 2.0 license.}

% TODO(rws). Note somewhere also that the error rates for English are down in the noise of interannotator agreement.

\section{Ranking models}
\label{sec:ranking}

% TODO(wuke): Describe MaxEnt ranking; define conditional set.
% TODO(wuke): Describe how we align spoken tokens and written tokens.

The discussion in this section is applicable to both the language specific normalizer grammar and the covering grammar, and so we refer to both as simply \emph{the grammar}.

For a written token $x_i$ in sentence $\mathbf{x}$, the set of possible outputs $Y_i$ consists of both the result of composition of $x_i$ with the grammar as well as $x_i$ itself (which corresponds to passing the token through, unmodified).
When $|Y_i| > 1$---i.e., when there are multiple normalization options for $x_i$---we need some way to choose the contextually appropriate output $\mathbf{y}^*_i \in Y_i$ which is closest to the reference verbalization $\mathbf{z}_i$.
The simplest such model is an n-gram language model (LM) built from the spoken side of the parallel training data.\footnote{This is essentially the approach taken by \citealt{Sproat2001}.}
We refer to this as the \emph{baseline system}.
% mention Sproat et al. 2001 at some point? --kbg
% done
However, there are two limitations with this approach.
First, the vast majority of the LM's parameters are otiose as they pertain to n-grams unaffected by the normalization process.
Secondly, the LM scoring makes no use of knowledge about the written inputs or about the grammar.

We therefore cast the choice of $\mathbf{y}_{i,j}$ as a ranking problem.
For each $Y_i$, let $G_i = \{\mathbf{y}\ |\ d(\mathbf{y}, \mathbf{z}_i) = \min_{\mathbf{y}'} d(\mathbf{y}', \mathbf{z}_i)\}$ be the subset of \emph{good} candidates from $Y_i$, according to some distance metric $d(\cdot)$ such as label edit distance.
Given a feature function $\Phi(\mathbf{y}_{i,j}, \ldots)$ that combines the candidate verbalization and properties of the context into a feature vector, we train a maximum entropy ranker \cite{sproat2014applications} by choosing $\mathbf{w}$ that maximizes
\[
  L(\mathbf{w}) = \sum_i \log \frac{\sum_{\mathbf{y}_{i,j} \in G_i} \exp \langle \mathbf{w}, \Phi(\mathbf{y}_{i,j}, \ldots) \rangle}{\sum_{\mathbf{y}_{i,j} \in Y_i} \exp \langle \mathbf{w}, \Phi(\mathbf{y}_{i,j}, \ldots) \rangle}
\]
The exact training and inference algorithms depend on the choice of feature function.
In this paper, we experiment with two classes of feature functions.

\subsection{Local ranking}
\label{subsec:local_ranking}

We can use a local feature function $\Phi_I(\mathbf{X}, i, \mathbf{y})$, which sees the whole input and the current verbalization, but not the verbalization of any other written tokens.
Such feature functions allow independent inference for each written token.
For training, we simply generate one training example for each $x_i$ for which $|Y_i| > 1$.
We use the following features in our experiments:

% Can this be compressed vertically? Right now it takes up a lot of space --KBG
\begin{compactdesc}
  \item[Local output n-grams:] n-grams (with $n = 1, 2, 3$) within output $y_i$;
  \item[Boundary trigrams:] two written words on the left $(x_{i-2}, x_{i-1})$ and the first word of $y_i$; two written words on the right $(x_{i+1}, x_{i+2})$ and the last word of $y_i$;
  \item[Written/spoken skip-grams:] pairs of one written word on either the left or the right within a 4-word window and an output word in $y_i$;
  \item[Bias:] $\mathbf{1}_{x_i=y_i}$, i.e., whether $x_i$ is passed through.
\end{compactdesc}

\subsection{Discriminative language model}
\label{subsec:dislm}
In normalization, the majority of the written tokens are actually passed through, therefore a feature function $\Phi_O(\mathbf{y}_{1,j_1}, \ldots, \mathbf{y}_{i,j_i}, \mathbf{1}_{x_i=\mathbf{y}_{i,j_i}})$ that sees the verbalization history but nothing much about the written tokens actually has a lot of overlapping information with $\Phi_I(\cdot)$.
We therefore limit the features to spoken token n-gram suffixes ending at $\mathbf{y}_{i,j}$ and the bias $\mathbf{1}_{x_i=\mathbf{y}_{i,j}}$. %so that %rws this is an orphan or did you intend to complete this sentence with something else?
Further, we fix the weight of the bias feature to a constant negative number because passing through should be discouraged, and leave the n-gram weights as the only tunable parameters.
Then, the trained model can be encoded as a WFST \cite{wu2014encoding} for efficient inference.

This feature parameterization also allows us to train a model from spoken tokens without any information about the written tokens. In effect we are ``hallucinating'' the written side in a way similar to \citealt{Sagae:EtAl:12}.
Consider the following simple training procedure with $\Phi_O(\cdot)$: for each $Y_i$, we extract feature vectors by assuming all the previous verbalizations are correct, i.e., $\Phi_O(\mathbf{z}_1, \ldots, \mathbf{z}_{i-1}, \mathbf{y}_{i,j}, \mathbf{1}_{x_i=\mathbf{y}_{i,j}})$, and train $\mathbf{w}$ to assign higher scores to those $\mathbf{y}_{i,j} \in G_i$.
If we only have spoken tokens $\mathbf{Y}$ but not the corresponding written tokens, we can approximate the above example generation for any sub-sequence $\mathbf{y}_i$ in $\mathbf{Y}$ by approximating $Y_i$ by $Y'_i = \pi_o(\mathbf{y}_i \circ V^{-1} \circ V)$, where $V$ is the WFST representing the grammar, and $G_i$ by $G'_i = \{\mathbf{y}_i\}$.
The output of the composition $\mathbf{y}_i \circ V^{-1}$ is the set of written tokens from which the grammar may produce $\mathbf{y}_i$.
Then, the cascaded composition gives all the spoken tokens the grammar may produce from a written token that produces $\mathbf{y}_i$.
If $V$ has perfect coverage over the written tokens, then $Y'_i$ is a superset of $Y_i$ and $G'_i$ is a subset of $G_i$.
To control the amount data generated, we only allow $\mathbf{y}_i$ to be at most 5 words, and limit the size of $Y'_i$ to be the 10,000 shortest paths through the cascade.
For example, the spoken sequence \emph{one twenty} might be mapped via $V^{-1}$ to various written forms such as \emph{120, 1:20, one20}, inter alia. These in turn composed with $V$ would yield various potential spoken forms, including \emph{one twenty, one hundred twenty, twenty past one}, inter alia.

% TODO: decide if we want to say something along the following lines about the MT experiments.
% In early work on this problem we also experimented with an MT system ... While the overall performance was comparable (XXX WER), it was particularly bad at learning certain classes of normalizations, such as numbers, producing incorrect results in many cases. Note that using any of the grammar-based approaches we do describe, this result is impossible: a properly constructed normalization grammar or normalization covering grammar cannot produce an impossible form for a number.  For reasons of space we do not report on this here.
% rws: I'd suggest not worth it

% wuke: This error rate is num of errors / num of total spoken tokens, which is hard to interpret because the reader does not know how many spoken tokens are actually normalized tokens. Should we change the denominator and use "normalized" error rate instead?  
% rws: yes I guess that makes sense.

\subsection{Candidate pruning}

Although the number of normalization candidates for a single written token almost never exceeds 100 for English, the number of candidates in Russian can be prohibitively large.
This is because the Russian number grammar permits all possible morphological variants for each output word, even though most combinations are ill-formed, leading to a combinatorial explosion.
We therefore bias the output by composing the number grammar with a local language model over the spoken (output) side of the number grammar.
The language model was trained using Witten-Bell smoothing and a held-out corpus of approximately 10,000 randomly-generated numbers.
We note it is also possible to mine this data from the web by identifying strings which match the output projection of the unbiased grammar \cite{Sproat2010}.

\subsection{Discriminative LM vs local ranking}
\label{sec:full-dis-lm}

As we will see, there is a large gap in error rates between discriminative LMs trained on real data and local ranking, especially with language-specific grammars.
This is due to the lack of two pieces of information not easily available for hallucinated data, the more interesting use case of the discriminative LM method:
\begin{compactdesc}
\item[Tuned pass-through bias] We fix the bias because we cannot tune it with hallucinated data.
However, the grammar sometimes produces incorrect verbalizations for common words that should really be passed through, often because of pecularities in the data annotation.
When training on real data, we can easily fix this issue by tuning the bias along with n-gram weights.
\item[Spoken phrase boundary] Consider the input \emph{1911 9mm}, whose correct verbalization is \emph{nineteen eleven nine millimeter}.
The discriminative LM prefers \emph{one nine one one nine millimeter}, because the hallucinated written form includes \emph{19119 mm}, and there is a strong bias in these data to reading five-digit numbers as digit sequences.
With real data, we know where the spoken phrase for a single written word begins or ends, and can train discriminative LMs on data with a special marker \emph{<p>} inserted at such boundaries.
This way, the LM can distinguish between \emph{one nine one one <p> nine} and \emph{one nine one one nine}, and know that the former is less likely than \emph{nineteen eleven <p> nine}.
\end{compactdesc}

\section{Results}
\label{sec:results}

We evaluate the text normalization systems using the following two metrics,
\begin{compactdesc}
  \item[Word Error Rate (WER)] The edit distance between the system output and the reference output, divided by the number of spoken tokens in the reference \emph{that are a result of normalization};
  \item[Sentence Error Rate (SER)] The proportion of test sentences with at least one error in the output.
\end{compactdesc}

\begin{table*}[t]
  %\label{tab:results_en}
  \centering
  \begin{tabular}{l c rrrr}
    \toprule
                               && \multicolumn{2}{c}{Language-specific} & \multicolumn{2}{c}{Covering}\\
    System                     && WER     & SER     & WER     & SER\\
    \midrule
    \\
    \textbf{English} \\
    \cmidrule{1-1}
    Baseline                   && 10.02\% & 14.14\% & 13.65\% & 20.06\% \\
    Local ranking              &&  7.52\% & 10.38\% & 10.14\% & 13.40\% \\
    Dis. LM, real data         &&  8.59\% & 11.90\% & 10.57\% & 14.06\% \\
    Dis. LM, real data, +tuned bias &&  8.14\% & 11.18\% & 10.36\% & 13.82\% \\
    Dis. LM, real data, +tuned bias, spoken phrase boundary &&  7.50\% & 10.38\% & 10.49\% & 13.82\% \\
    Dis. LM, hallucinated data &&  8.80\% & 12.06\% & 10.92\% & 14.52\% \\
%    \bottomrule
%  \end{tabular}
%  \caption{Error rate results for English}
%\end{table*}
%
%\begin{table*}
%  \label{tab:results_ru}
%  \centering
%  \begin{tabular}{l c rrrr}
%    \toprule
%                               && \multicolumn{2}{c}{Language-specific} & \multicolumn{2}{c}{Covering}\\
%    System                     && WER     & SER     & WER     & SER\\
%    \midrule
\\
    \textbf{Russian} \\
    \cmidrule{1-1}
    Baseline                   && 22.29\% & 25.45\% & 26.05\% & 35.25\% \\
    Local ranking              && 14.85\% & 19.85\% & 17.66\% & 24.50\% \\
    Dis. LM, real data         && 18.79\% & 23.75\% & 17.79\% & 23.45\% \\
    Dis. LM, real data, +tuned bias &&  15.54\% & 20.25\% & 17.70\% & 23.35\% \\
    Dis. LM, real data, +tuned bias, spoken phrase boundary &&  15.28\% & 19.90\% & 17.47\% & 23.10\% \\
    Dis. LM, hallucinated data && 18.44\% & 22.30\% & 19.67\% & 26.05\% \\
    \bottomrule
  \end{tabular}
  %\caption{Error rate results for Russian}
  \caption{Error rate results for English and Russian.}
  \label{tab:results}
\end{table*}

For the baseline systems, we use unpruned trigram\footnote{Using higher order n-grams does not lower WER on the development set.} language models trained on the spoken side of the training data, with Katz smoothing.
We also experiment with the variants of discriminative LM discussed in Section \ref{sec:full-dis-lm}, by adding \emph{tuned bias} and \emph{spoken phrase boundary} when training on real data.
For all the systems, we set hyperparameters so as to minimize WER over the development set.

Results for English and Russian are shown in Table~\ref{tab:results}.\footnote{\label{fn:tts}While the error rates of even the best systems in Table~\ref{tab:results} may seem high, in early experiments with these same data we found that the methods here actually outperformed those of a commercial TTS normalization system \cite{Ebden:Sproat:14}. This is largely due to the fact that the TTS normalization system was tuned for different data from the voice search domain so that, for example a digit sequence like \emph{920} has a default reading in the TTS system as \emph{nine hundred twenty}, whereas in the voice search domain it is much more likely to be \emph{nine twenty}.}
We observe that all of the ranking-based systems outperform the baseline system that uses the same grammar. Indeed in Russian, all ranking-based systems outperform both baselines regardless of the grammar used.
The cost of using the covering grammar as opposed to the language-specific grammar is about a 25\% relative increase in WER for English, but only about a 6\% increase for Russian, suggesting that the hand-built Russian grammar has been less carefully curated than the English grammar.
Local ranking with all the features discussed in Section~\ref{subsec:local_ranking} yields the best results, with a penalty for the discriminative LM (Section~\ref{subsec:dislm}) with a language-specific grammar of between 14\% relative (English) to 25\% relative (Russian) WER. When using a covering grammar, the discriminative LM and local ranking are much closer to each other in performance.
Finally, without tuned bias or spoken phrase boundary, the difference between the discriminative LM trained on real written-spoken correspondences, and that trained on hallucinated data is usually quite small ($<$1\% absolute), except in the case of the Russian covering grammar.
However, with the language-specific grammar, the hallucinated training scenario actually performs slightly \emph{better} than training with real written-spoken correspondences. 
Note also that with real data the covering grammar outperforms the language-specific grammar for Russian with the discriminative language model.

As is obvious from Table~\ref{tab:results}, Russian has much higher error rates overall than English, reflecting the greater morphological complexity of this language, as well as other things such as the fact that in these data, tokens written as digit sequences could be read as either cardinal or ordinal numbers in Russian (but only rarely as ordinals in English), and thus the system must resolve that ambiguity as well as handling case, gender and number morphology.

In summary, local ranking solutions outperform the baseline spoken-side trigram LM as a method to rank the candidates generated by the grammar. Local ranking outperforms discriminative language model ranking without tuned bias or phrase boundary information, but controlling for the same setup in discriminative language model experiments, hallucinated data does not hurt performance by very much.
These results suggest that one can develop an initial system with reasonable performance for a new language with very little work if one just has a source of ``spoken'' text, and is willing to invest a small amount of effort in developing a list of lexical entries, as well as some example number names --- cf. \cite{Gorman:Sproat:16} --- for the language. 
Hallucinated data already approximates the ambiguity a ranking model needs to resolve very well and future effort should focus on approximating pass-through bias and phrase boundary in discriminative LM training.
Naturally one can do better if one develops more language-specific grammars, as well as aligned written and spoken data.
But the lowest-resource scenarios may be sufficient for an initial system, and other---higher-resource--- scenarios we propose suggest a path one might take to improve such a system.

\section{Qualitative error analysis}
\label{sec:error_analysis}

We performed a qualitative error analysis in order to see if there were any broad generalizations about the kinds of errors that could be attributed to using the covering grammar versus the language specific grammar, and hallucinated versus real training data. We present these results below.

\subsection{English errors}

For the covering grammar (vs language-specific grammar), the main error categories were the following:
\begin{compactitem}
\item Derived numerical expressions like \emph{49ers}, which the covering grammar does not support.
\item Occasional readings of words letter-by-letter:  \emph{s o m e t h i n g}.
\item Explicit \emph{o'clock} in times (e.g. \emph{three o'clock p m} versus \emph{three p m}).
\item Digit readings of some numbers, such as \emph{6308} as \emph{six three oh eight} instead of \emph{sixty three oh eight}.
\item Hundreds readings in cases like \emph{2200}---\emph{two thousand two hundred} versus \emph{twenty two hundred}.
\end{compactitem}
Many of these categories of errors can be attributed to language-particular constructions that are simply not handled by the covering grammars, e.g., the case of \emph{49ers}.

For the hallucinated versus real data the main categories were, again, digit-by-digit readings of some numbers; and reading of \emph{+} as \emph{and} rather than \emph{plus}.  This second error is presumably due to the fact that \emph{+} may be read as \emph{and}, and in the construction $V^{-1} \circ V$, we map from \emph{and} to \emph{+} and back to \emph{and} and \emph{plus} as candidates. As \emph{and} is far more common that \emph{plus}, the system learns to prefer \emph{and} as a reading of \emph{+}.

\subsection{Russian errors}

The Russian errors for covering grammars vs. language-specific grammars fall mostly into the following categories:
\begin{compactitem}
\item Failure to read точка `dot' in URLs.
\item Use of cardinal numbers when ordinals would be more appropriate: пятнадцать августа `August 15' rather than  пятнадцатого августа.
\item A small number of instances of the special case of using раз for `one' in counting (раз два три --- 1 2 3), which the number grammar does not permit.
\end{compactitem}
The case of точка `dot' is apparently due to an omission in the covering grammar, which does not handle the \texttt{ru} top-level domain.
For hallucinated versus real data, the main errors were:
\begin{compactitem}
\item Reading words as letter sequences or vice versa:  а к сорок семь `AK-47' read as ак сорок семь (i.e. `ak forty seven' rather than `a k forty seven')
\item Digit-by digit readings versus grouped readings: \emph{4400} as четыре четыре ноль ноль (`four four zero zero') versus сорок четыре ноль ноль (`forty four zero zero').
\end{compactitem}

\section{Discussion and Future Work}
\label{sec:discussion}

We have explored the relative contribution of different hand-built data to the performance of text-normalization systems trained using ranking. Specifically:
\begin{compactitem}
\item language-specific normalizer grammars versus language-independent covering grammars with a small amount of language-specific lexical knowledge; and 
\item aligned written-spoken data versus only spoken data ``hallucinating'' the written side.
\end{compactitem}
We have shown that the performance degradation for using a covering grammar over a language-specific grammar need not be large---and in one configuration for Russian, the former actually outperformed the latter. In a similar fashion, the degradation caused by hallucinated data need also not be large; again in Russian, the hallucinated scenario slightly outperforms the discriminative LM trained on real data.

The choice of English and Russian in these experiments was motivated by the fact that we already had text normalization systems for these languages (see Footnote~\ref{fn:tts}), and we could thus compare directly with existing approaches.  But of course the real interest in these methods is in developing systems for new languages, and in particular low-resource languages where we can get raw data, and some linguistic knowledge, but do not have the resources to invest in large-scale grammar development. Therefore we plan to apply the methods reported here to low-resource languages in future work.

\section*{Acknowledgments}

We thank Michael Riley for much discussion of various versions of this work.

\bibliography{refs}
\bibliographystyle{acl}

\end{document}